\documentclass[11pt]{article}
\usepackage{coling2018}
\usepackage{times}
\usepackage{latexsym}
\usepackage{multirow}
\usepackage{url}
\usepackage{graphicx, dblfloatfix}
\usepackage{algorithm}
\usepackage{algorithmic}
\usepackage{svg}

\title{Towards Understanding End-of-trip Instructions in a Taxi Ride Scenario}

\author{Deepthi Karkada$^{2*}$, Ramesh Manuvinakurike$^{1*}$, Kallirroi Georgila$^1$ \\
$^1$Institute for Creative Technologies, University of Southern California \\
  $^2$Intel Corp\\
  {\tt deepthi.karkada@intel.com, [manuvinakurike,kgeorgila]@ict.usc.edu} \\}

\date{}

\begin{document}
\maketitle
\begin{abstract}
\blfootnote{$^*$ Equal contribution.}
We introduce a dataset containing human-authored descriptions of target locations in an ``end-of-trip in a taxi ride'' scenario. 
We describe our data collection method and a novel annotation scheme that supports understanding of such descriptions of target locations. 
Our dataset contains 
target location descriptions 
for 
both synthetic and real-world images as well as visual annotations (ground truth labels, dimensions of vehicles and objects, coordinates of the target location, distance and direction of the target location from vehicles and objects) that can be used in various visual and language tasks. 
We also perform a pilot experiment on how the corpus could be applied to visual reference resolution in this domain. 
\end{abstract}

\section{Introduction}

The last few utterances in a typical taxi ride are the passengers directing the driver to stop their ride at the desired target location. ``Stop right next to the white car'', ``behind the big tree should work'', ``drop me off in front of the second black pickup truck'' are all examples of such utterances. 
Resolving these requests, while a simple task for the human drivers, assumes complex vision and language understanding capabilities. 
Some of the sub-tasks that the driver needs to perform to resolve these requests are: i) Visual reference resolution: Identifying the visual objects that the rider is referring to (called the referent(s)) among the visual distractors present in the scene (the big tree, the second black pickup truck, the white car, etc.); ii) Directional description understanding: Predicting the target location that the rider refers to with respect to the referent(s) present around (in front of, right behind, a little further from, etc.); and iii) Action identification: The action that the rider wants to take (stop, drop me off, etc.). The purpose of this work is to build a dataset that comprises of such utterances and build an annotation scheme supporting the understanding of such utterances. 

\begin{figure}[!ht]
    \centering
    \includegraphics[width=0.65\columnwidth]{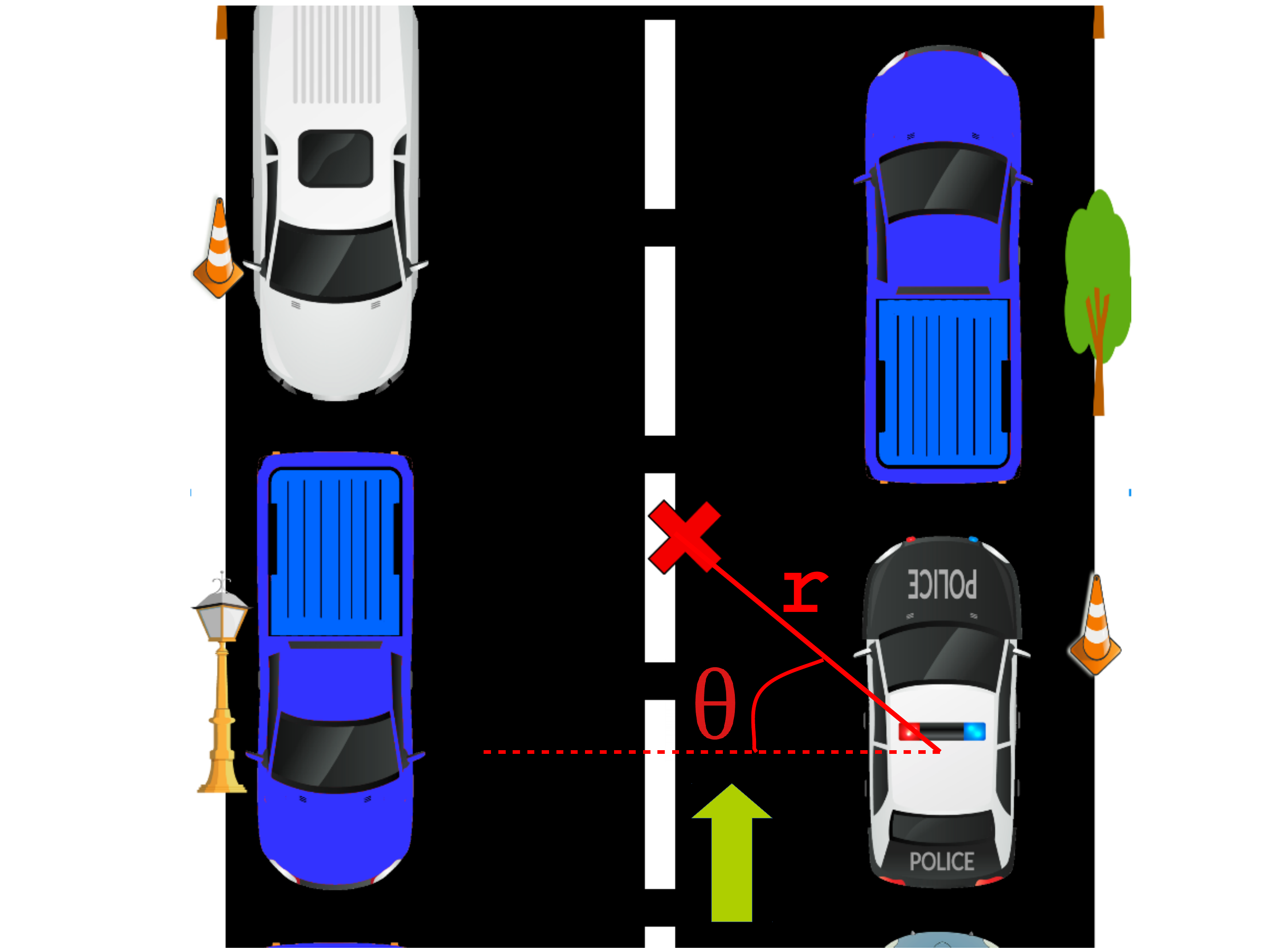}
    \caption{Example from the synthetic section of the dataset. The annotation labels of r and $\theta$ define the target location for an example utterance ``drop me off in front of the cop car''. The green arrow shows the direction of motion of the taxi.}
    \label{fig:rtheta}
  \end{figure}

We introduce a novel dataset which contains the human-authored natural language descriptions of the desired target location in an end-of-trip taxi ride scenario with synthetic images and real street images. 
We describe the annotation scheme for these descriptions which comprises of referents, directional descriptions, and actions, and show that the inter-annotator agreement is high. 
Our dataset contains the images with the ground-truth target location coordinates that are described by the users. 
The image annotations also contain object ground-truth labels, coordinates, dimensions along with the distance and direction of the target location with respect to the objects that are present in the image. 
We refer to the position of the target location as a function of `r' and `$\theta$' where `r' is the magnitude of the vector, and $\theta$ is the direction between the referent and the target location. 
This quantification provides the capability to predict the target location coordinates using natural language sentences given the visual context. 
Figure~\ref{fig:rtheta} shows an example where the combination of r and $\theta$ determines the target location with respect to the referent(s). 

The contributions of this work are: 1) A novel corpus containing user descriptions of target locations for synthetic and real-world street images.
2) The natural language description annotations along with the visual annotations for the task of target location prediction. 
3) A baseline model for the task of identification of referents from user descriptions.

\section{Related Work}

There is a strong relation between the language and vision modalities, and 
the information in the vision modality influences the associated spoken language \cite{tanenhaus1995integration}. 
In recent times, automating various tasks involving vision and language has attracted much interest. The task of reference resolution is one such example. This task typically involves identification of one of the objects referred to in a set of similar distractors through dialogue \cite{clark1986referring,kennington2015simple,maikeeveagent,de2017guesswhat,manuvinakurike2017using}. 

Other tasks that combine language and vision are:
visual question answering which requires answering questions about an image \cite{antol2015vqa}, a related question generation task \cite{mostafazadeh2016}, 
storytelling \cite{visualstorytelling2016}, and conversational image editing \cite{manuvina-editme,manuvinakurike2018sigdial}.
Furthermore, other relevant approaches are automatic image captioning and retrieval by using neural networks to map the image into a dense vector, and then conditioning a neural language model on this vector to produce an output string  \cite{mitchell-et-al:2012,kulkarni2013babytalk,socher2014grounded,vinyals2015show,devlin2015language}. 

Annotation of spatial information including objects and their spatial relations in real-world images has been studied in detail for developing the ISO-Space annotation scheme \cite{pustejovsky2011using,pustejovsky2014image}. The semantics of spatial language have also been studied in detail; see for example \newcite{varzi2007spatial} and \newcite{bateman2010linguistic}. The focus of our work is not on the study of spatial semantics but rather on the task of target location identification using simplistic annotations.

The goal of this work is to study user descriptions in an ``end-of-taxi'' ride scenario which involves studying language and vision in a situated environment. 
Related to our work, 
\newcite{lemon2006isu} built a dialogue system for an in-car domain and
\newcite{eric2017key} studied dialogues with regard to helping a driver navigate to a specific location. However, these works did not specifically study the interaction and combination of the vision and language modalities in a situated in-car environment. 
Our work contributes to the literature with a corpus combining the language and vision modalities in a situated environment. 
We extract the embedded representations of descriptions generated from the users and use them for the task of reference resolution by comparing them to similar embeddings extracted for the object ground truth labels. 
We also discuss r$\theta$ annotations that can be used to understand directional relations using the outputs of the reference resolution module, which is a particularly novel feature of our annotation scheme. 
Note that in prior work,
reference resolution is performed using models that understand the meaning of words using classifiers trained with visual features \cite{kennington2015simple,manuvinakurike2016real}.

\section{Data Collection}
\label{sec:datacollection}

\begin{figure}[!ht]
    \begin{center}
    \includegraphics[width=\columnwidth]{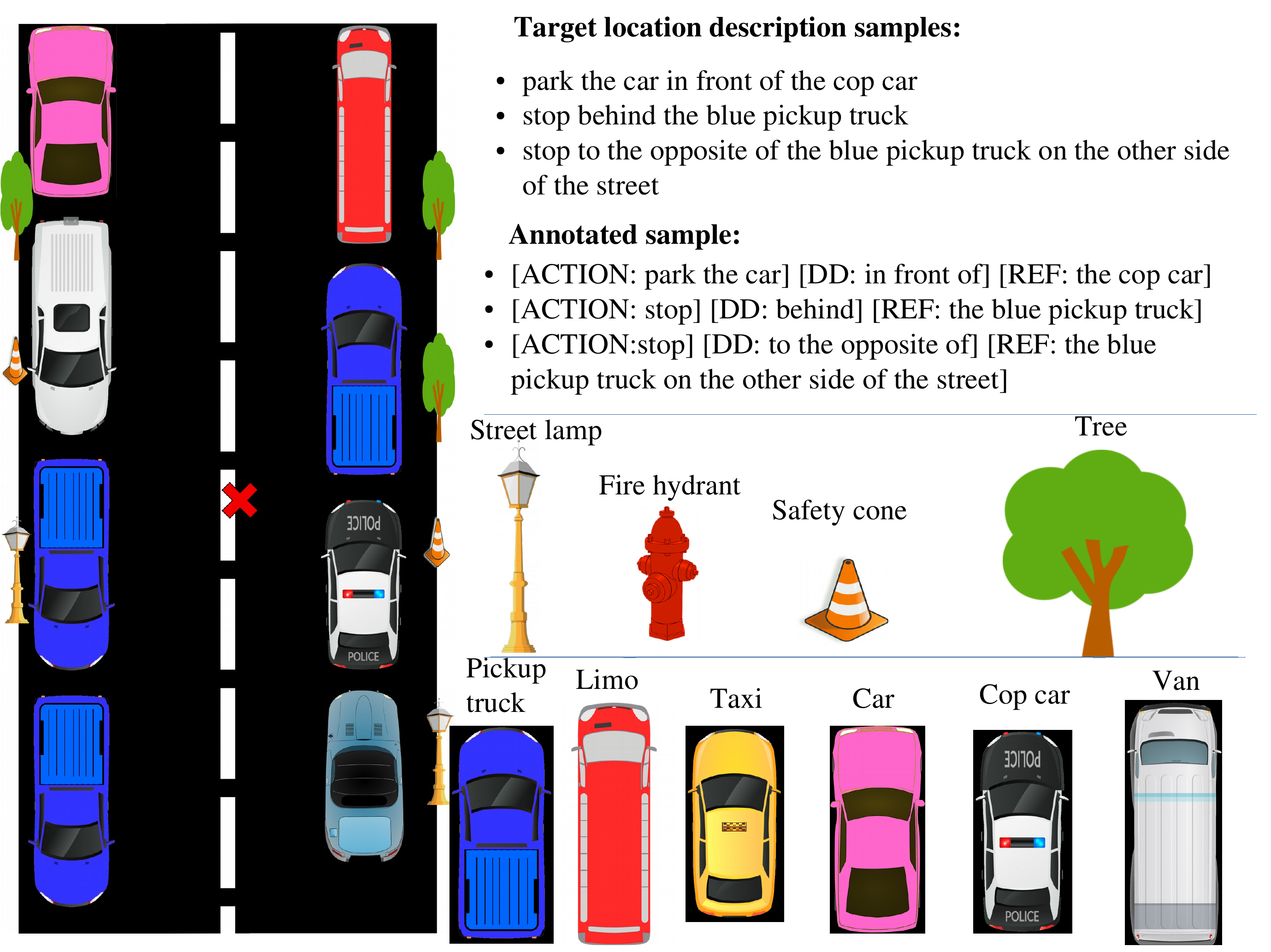} 
    \caption{Sample data from the synthetic image set. The figure shows sample user instructions and the annotations performed. The ground truth labels of the vehicles and objects are also provided. The color information of the vehicles is also present in the dataset. }
    \label{fig:twodex}
    \end{center}
  \end{figure}

We use the crowd-sourcing paradigm\footnote{https://www.mturk.com} to collect user descriptions 
instructing a taxi to stop at a given location (we will refer to this location as the ``target location''). 
The Amazon Mechanical Turk users (called turkers) are shown an image (similar to Figure~\ref{fig:twodex} or Figure~\ref{fig:threedex}) and are asked to imagine a scenario where they are in a taxi about to reach their destination. 
As they approach their destination 
they need to instruct the taxi driver in natural language to stop at the bright red cross. 
The turkers needed to provide at least three unique descriptions. Only native English speakers whose location was the US (United States) were chosen for the task. 

The images shown to the turkers contain vehicles and other objects which are used as referents to describe the target location. 
These images were either i) Synthetic (programmatically generated) or ii)  Street-view  images (extracted from Google maps\footnote{https://maps.google.com}) which we refer to as real-world images. 
The synthetic images are generated with a 2-dimensional top-view perspective containing vehicles that are typically observed on the streets in US parked on either side of the street. 
The street-view images are collected using a popular street navigation application (Google street-view\footnote{https://maps.google.com, official endorsement should not be inferred.}), which contains the images of real streets taken from a car-top mounted camera. 
Below we describe the methods followed in the construction of these images. 

\subsection{Synthetic Images Construction}
\label{subsec:syn}

The synthetic images were constructed programmatically by overlaying the vehicle and object templates on the street with a median in the middle. 
The synthetic images were constructed from bird's eye point of view which helps us overcome the problem of frame of reference. 
Templates of the different categories of vehicles such as cars (of different colors\footnote{The colors of the cars were chosen based on the most common car colors. We chose blue, brown, green, grey, yellow, orange, pink, red, and white. We did not choose black as it is difficult to spot against the background.}) including taxi and police cars, pickup trucks\footnote{Blue and white color.}, red truck, white van, and limousine were overlaid on either side of the street randomly. 
The vehicles were placed in a parked position on a two-way right-side-driving street\footnote{As the majority of countries are right-side-driving, we choose the right-side-driving orientation for generating the images.}. 
Four objects (street lamp, fire pump, traffic cone, and tree) were placed on the sidewalk randomly.
A maximum of up to 4 vehicles were placed on either side of the street. The distance between the vehicles was not uniform. 
Figure~\ref{fig:twodex} shows a sample synthetic image: vehicles and objects along with three user-authored descriptions. 
A ``red cross'' was also randomly placed on the street part of the image which was to be used by the users as the target location for the taxi to stop. 
The synthetic images provide an environment devoid of complexities (e.g., visual segmentation, object identification, masking) otherwise present in real-world images which are not the focus of this work. 



\subsection{Real-World Images Construction}

\label{subsec:rwd}
  \begin{figure}[!ht]
   \centering
    \includegraphics[width=0.85\columnwidth]{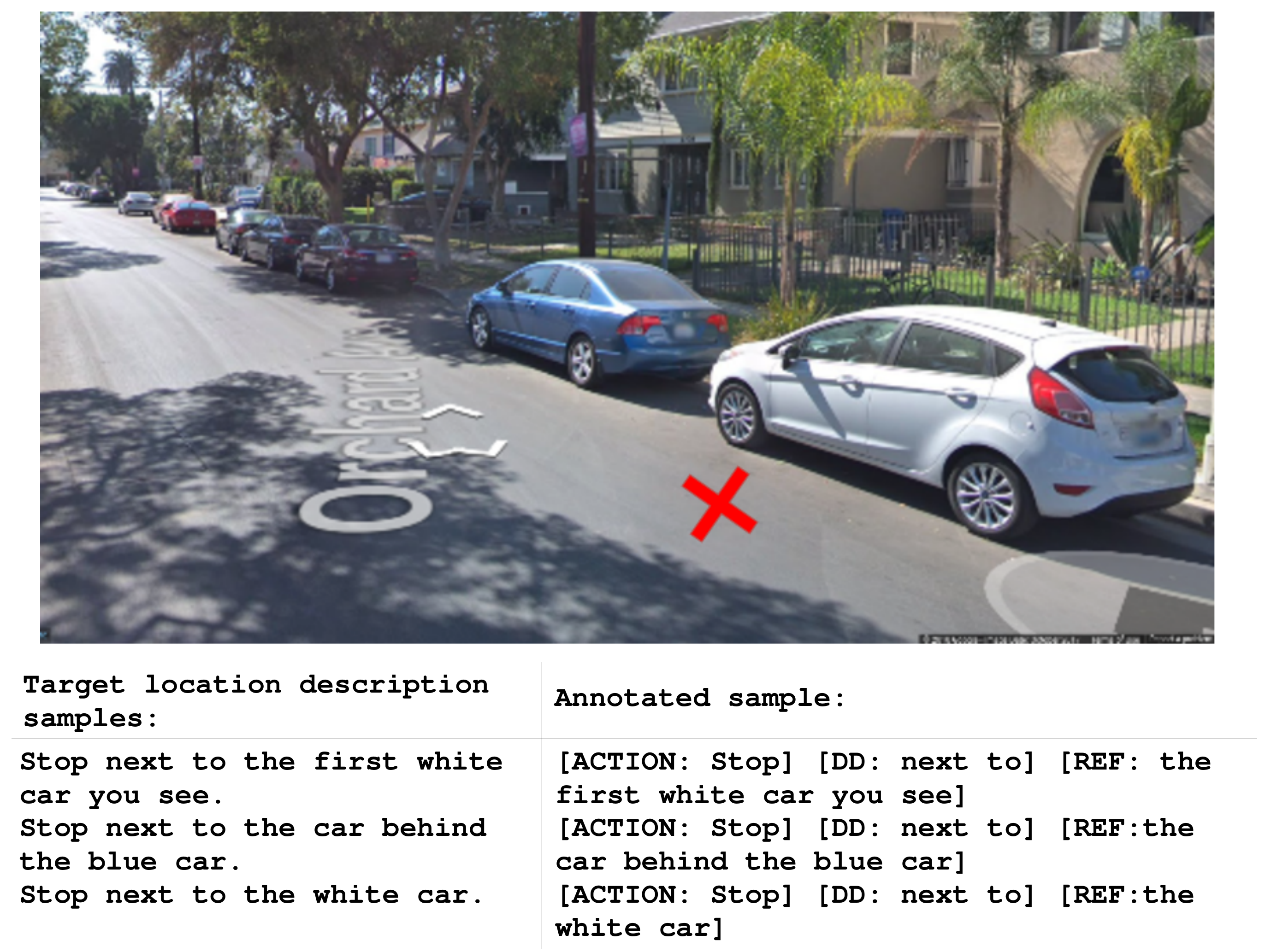}
    \caption{Example data from the real-world 3d street images.}
    \label{fig:threedex}
  \end{figure} 
 
We extracted the real-world images from Google street-view imagery in the streets of a busy city. 
The images were captured manually using the Linux snapshot tool. 
Since, the street-view images are taken from the roof-mounted camera placed on a right-side-driving vehicle we do not face the issue of unknown frame of reference. 
A sample image is shown in Figure~\ref{fig:threedex}. 
The ``red cross'' was overlaid randomly on the street which was the designated target location for the users to describe in the image.  


\begin{table}[!ht]
    \centering
    \begin{tabular}{| l | l | l | l | l | l | l | l |}
        \hline
        \multicolumn{3}{|c|}{Synthetic} & \multicolumn{3}{|c|}{Real-world} & \multicolumn{2}{|c|}{Combined} \\ \hline
        images & descriptions & tokens & images & descriptions & tokens & unique & common \\ \hline
         324 & 1069 & 9708 & 68 & 212 & 1863 & 457 & 128\\ \hline
        
    \end{tabular}
    
    \caption{\label{tab:data} Statistics of the dataset collected across the synthetic and the real-world images. The ``combined'' section contains the total unique tokens and common tokens shared across the descriptions in the synthetic and real-world. We observe that the difference in the language is mainly related to the referent descriptions.}
\end{table}

\subsection{Structure of the Descriptions}

Table~\ref{tab:data} shows the statistics of the dataset collected. 
The descriptions mainly consist of three parts: 
i) Actions: the words used by the user instructing the driver to perform an operation (e.g., ``stop'', ``keep going until''). 
Since we had directed the users to provide instructions for stopping, the actions for nearly all the actions specified were similar to the ``stop'' command. 
ii) Referent (REF): The words/phrases used to refer to the vehicles or the objects present in the image. 
The users typically refer to vehicles or objects close to the target location and these references are either simple or compound. 
In simple referent descriptions, the users refer to a single object in the scene, e.g., ``the blue car'', ``the white van''.  
In compound referent descriptions, the users refer to multiple objects such as ``the two blue vans'' in the phrase ``please park in front of the two blue vans''. They also use the position of the vehicles or objects such as ``the third car on the right'' in ``stop next to the third car on the right''. A few descriptions contained multiple referents, such as '`stop in between the taxi and the white SUV'`. In this case, we mark each referent separately. 
iii) Directional description (DD): This is the part of the description indicating direction that is used to refer to the target location in relation to the referent (REF). Instances of directional descriptions include phrases such as ``close to'', ``next to'', ``top of'', ``near'', ``between'', etc.  

Figures~\ref{fig:twodex} and~\ref{fig:threedex} show sample annotations. Two expert annotators annotated the same 25 randomly chosen descriptions to calculate inter-rater reliability. 
The annotations at the word level were considered to be the same if both the labels and the boundaries were agreed upon by both annotators.
The inter-rater reliability scores were measured using Cohen's kappa and was found to be 0.81 indicating high agreement. Most of the disagreements were limited to marking the beginning and the endpoints (typically articles and prepositions). 

  \begin{table}[t!]
    \centering
    \begin{tabular}{| l | l | l | l | l | l | }
        \hline
        \multicolumn{3}{|c|}{Synthetic} & \multicolumn{3}{|c|}{Real-world} \\ \hline
        Actions & Ref & DD & Actions & Ref & DD \\ \hline
        273 & 408 & 372 & 173 & 217 & 219 \\ \hline 
        
    \end{tabular}
    
    \begin{tabular}{| l | l | l | l | l | l | }
        \hline
        \multicolumn{3}{|c|}{Synthetic unique} & \multicolumn{3}{|c|}{Real-world unique} 
        \\ \hline
        Actions & Ref & DD & Actions & Ref & DD 
        \\ \hline
        8 & 185 & 89 & 13 & 181 & 75 
        \\ \hline 
        
    \end{tabular}
    
    \caption{\label{tab:data_annotations} Annotations statistics. }
\end{table}
We annotated a section of the data collected (see Table~\ref{tab:data_annotations}). 
We observed that there are fewer actions than user descriptions as a few turkers chose only to provide the directional description and referent. 
The number of referents and directional descriptions is greater than the number of total phrases. This is because the users provide compound descriptions mentioning multiple descriptions for the same target location (e.g., ``park to the left of the brown car, across the white van''). In such cases we label the referents and directional descriptions separately. 
There were also instances of images with multiple vehicles which looked similar.  
In such cases, the turkers supplemented the language used to identify the referent with descriptions of other objects. This can be observed in the description ``park the car near the blue sedan next to the light post'' where ``the blue sedan'' was not sufficient to identify the referent,  hence it was supplemented with further descriptions of the objects surrounding the referent. 
There are a lot more unique referent descriptions per unit description for the real-data as the array of real-world objects used for referents were more diverse.

  \begin{figure}[!ht]
    \begin{center}
    \includegraphics[width=0.49\columnwidth]{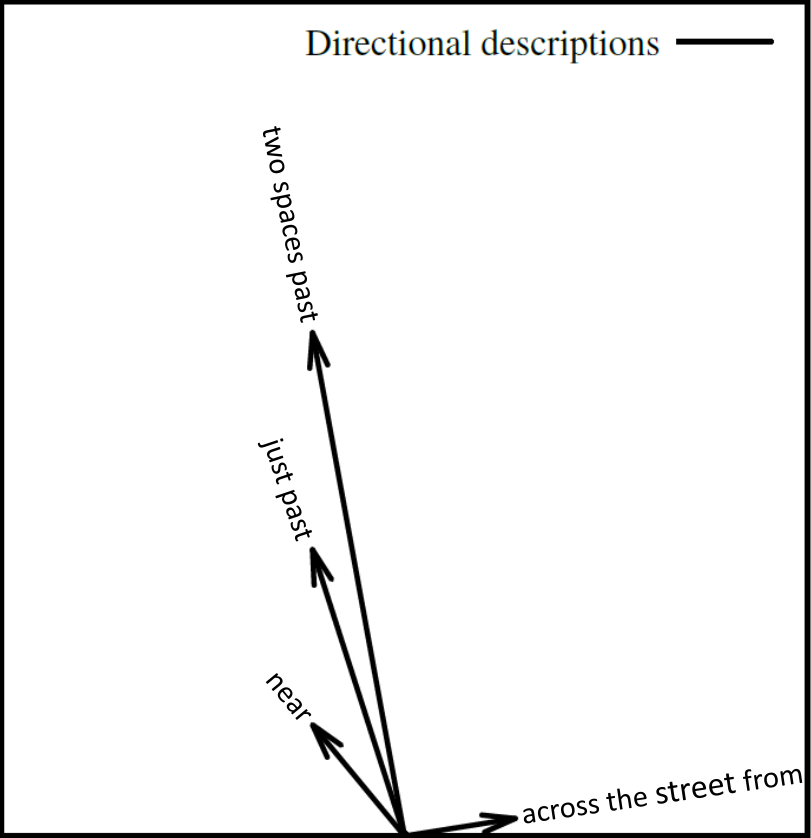}
    \includegraphics[width=0.49\columnwidth]{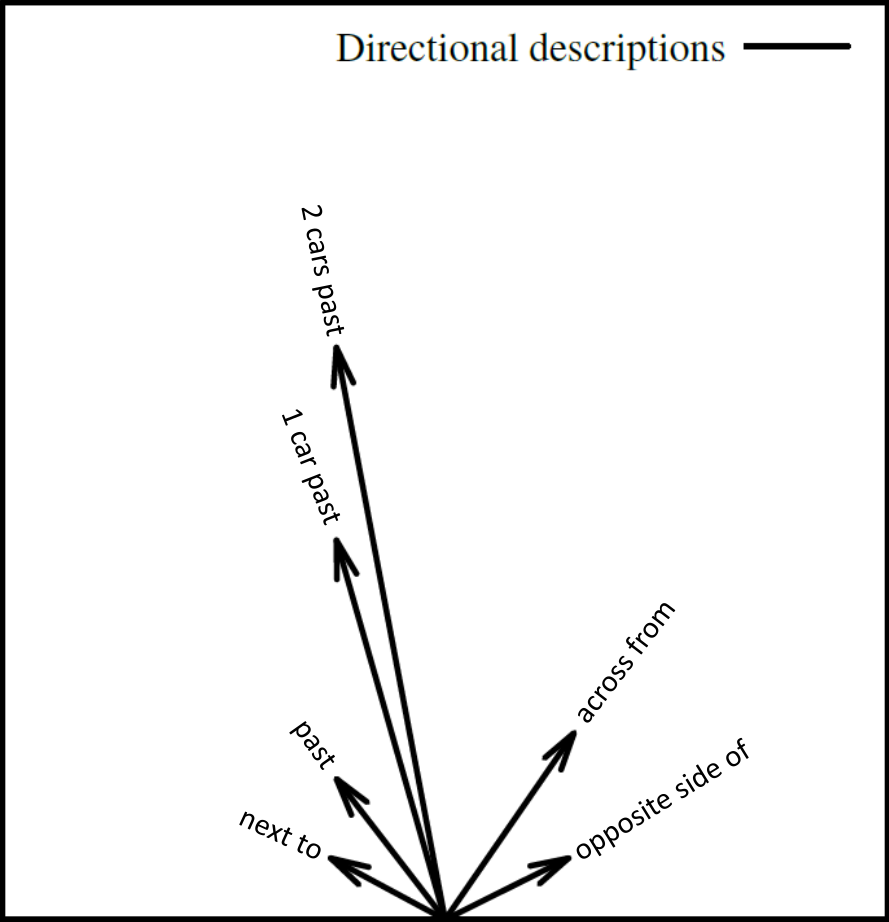}
    \includegraphics[width=0.49\columnwidth]{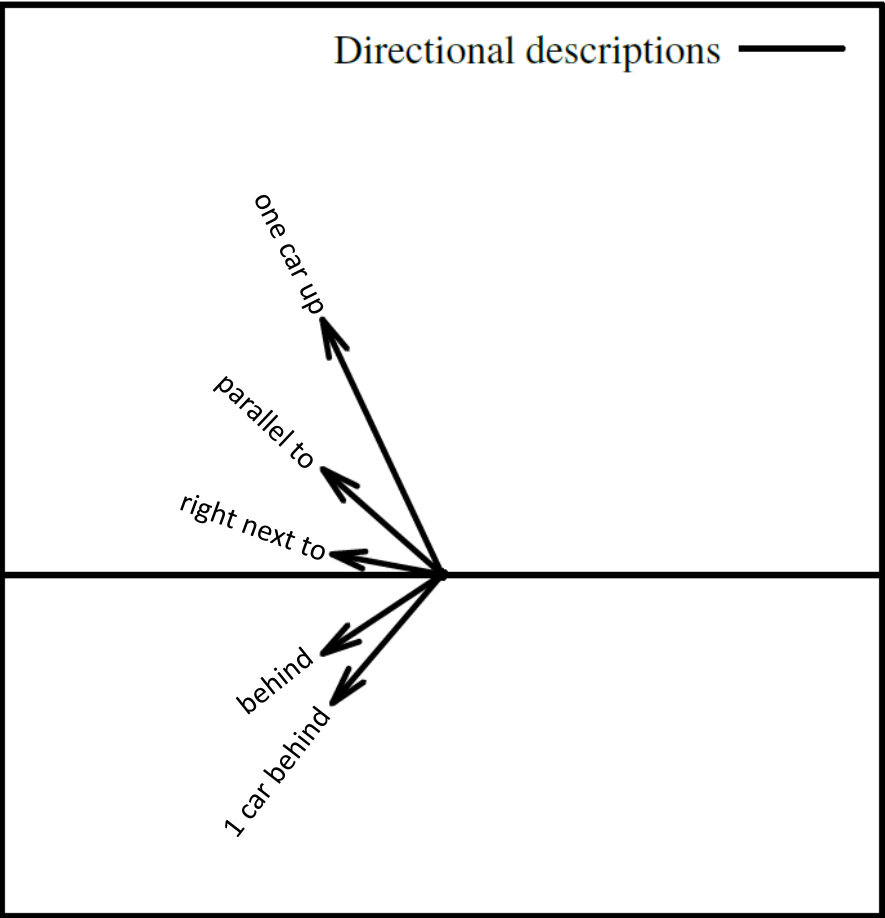} 
    \includegraphics[width=0.49\columnwidth]{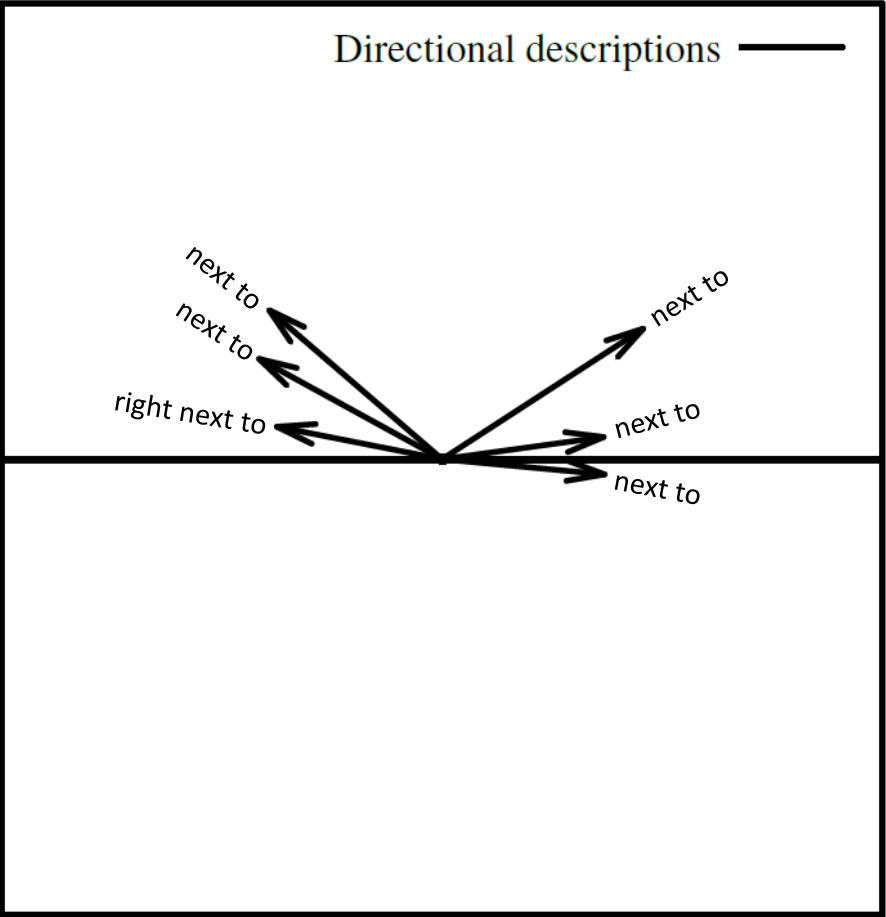}
    \caption{The graphs in the figure show the position of the target location (head of the arrow) with respect to the referent (nock of the arrow) for different images. The users describe the target location with respect to the referent. }
    \label{fig:dddesc}
    \end{center}
  \end{figure}  

\subsection{The r$\theta$  Labels} 

In order to estimate the coordinates of the target location, the coordinates of the objects in the scene, the directional description, and the referent are required. 
The dataset contains the coordinates of all the objects present in the image along with the ground truth labels.
Given the target location description, its position from the referent is available as an (r, $\theta$) tuple. 
Figure~\ref{fig:dddesc} shows a few examples. 
In Figure~\ref{fig:dddesc}, the position of the target location is shown with respect to the referent in different descriptions with the label of the directional description. 
The figure shows different directional descriptions from the referent to the target location.  
We can observe (top-left) that the directional description ``next to'' has a lower angle `$\theta$' and `r' (top left) compared to ``just past'' (as in ``stop just past the blue car'') which in turn has lower values of `$\theta$' and `r' compared to ``two spaces past''.  
We can also see (bottom-right) that ``next to'' is used to mean different positions with respect to the referent. 
``Behind'' typically refers to a negative value of `$\theta$' and ``right next to'' refers to a positive value (bottom-left). 
The synthetic and real-world data include 372 and 219 such directional descriptions respectively (see Table~\ref{tab:data_annotations}). 
The `r' in the r$\theta$  model refers to the radial distance between the center of the referent vehicle or object and the target location, and the `$\theta$' refers to the angle measured from the horizontal direction. 

\section{Understanding User Descriptions}

  \begin{figure}[t!]
   \centering
    \includegraphics[width=0.95\columnwidth]{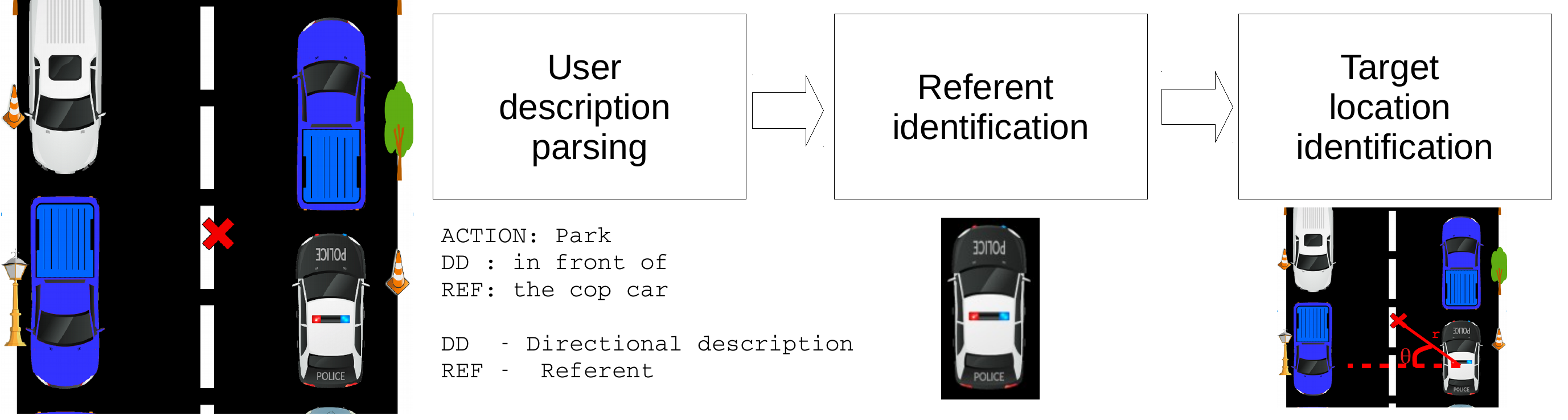}
    \caption{Task pipeline for identification of the target location using the user descriptions.}
    \label{fig:pipeline}
  \end{figure} 
  
Given the user description, images, and the annotations (language annotations such as action, referent, directional description, and visual annotations such as ground truth object labels and r$\theta$ labels), we define three separate tasks (see Figure~\ref{fig:pipeline}): 
i) Identification of the action, referent and directional relations descriptions in the user instructions: This step is also sometimes referred to as segmentation and intent labeling and is not the focus of this work. We assume oracle data, i.e., the complete and correct identification of the action, referent, and directional relations descriptions in the user instructions. 
ii) Referent(s) identification: The users refer to vehicles or objects in the images in their descriptions. The task of reference resolution refers to the identification of the correct vehicle or object based on the user description. In the cases where there are multiple referent descriptions, we identify all the referents in the dataset. 
iii) Target location identification: This task refers to combining the information from the referent identification and the directional relation description to identify the final target location. 

In this work we focus on the second problem. We do not perform parsing on the user descriptions and assume the availability of referent descriptions.  We will pursue the goal of automating the complete pipeline (see Figure~\ref{fig:pipeline}) in future work. Below we describe the referent identification task. 

\subsection{Referent Identification}

Given the referent description (REF) and the image, the task is to identify the object that is being referred to by the user. 
In this section, we describe the approach that we take to identify the referent based on the user description. 
We use the data for the synthetic images.
We assume the availability of the referent (text with REF label). 
The ground truth descriptions of the vehicles (e.g., pink car, white van, blue pickup truck) and the objects (e.g., fire pump, tree, traffic cone) are available from the image annotations (see Figure~\ref{fig:twodex}). 

The first approach is the ``random baseline''. Each image can have up to 16 vehicles and objects and randomly predicting one such object as the referent yields 6.25\% accuracy and is noted to be a random weak baseline. 
For the second approach we use the ``sub-string matching'' method to identify the referent object.
In this approach we compare the user provided referent string (text with REF label e.g., ``the pink sedan'') and the ground truth label (e.g., pink car, red car, white van) available from the images. 
We use the number of matching words to get the best match for a given image. In the case of a tie with multiple objects matching the same number of words, we randomly select one of the objects and check if the referent is correct. 
This method yields an accuracy of 47.5\%  which we use as a stronger logical baseline for comparison. 
This approach yields lower numbers because of the diverse set of vocabulary used to describe the referents. 
For instance, ``police car'' is referred to as ``cop car'' or ``sheriff's car''. 
To overcome this problem, we use the sentence embeddings approach \cite{mikolov2013distributed}. 

We obtain a vector representation of the referent description ($\vec{r}$) and the objects present in the image ($\vec{o}$). 
These vectors are generated using sent2vec \cite{pgj2017unsup} .
We then get the best candidate for the referent description by choosing the object with the maximum value of the dot product between the objects present in the image and the description. 
Thus the best suited object for the referent description is chosen using $\arg\max_{i} \vec{o}. \vec{r_i}$. 
The dot product is a measure of cosine similarity between the referent description (REF) and the ground truth labels. 

\begin{figure}[t!]
    \centering
    \includegraphics[width=0.25\columnwidth]{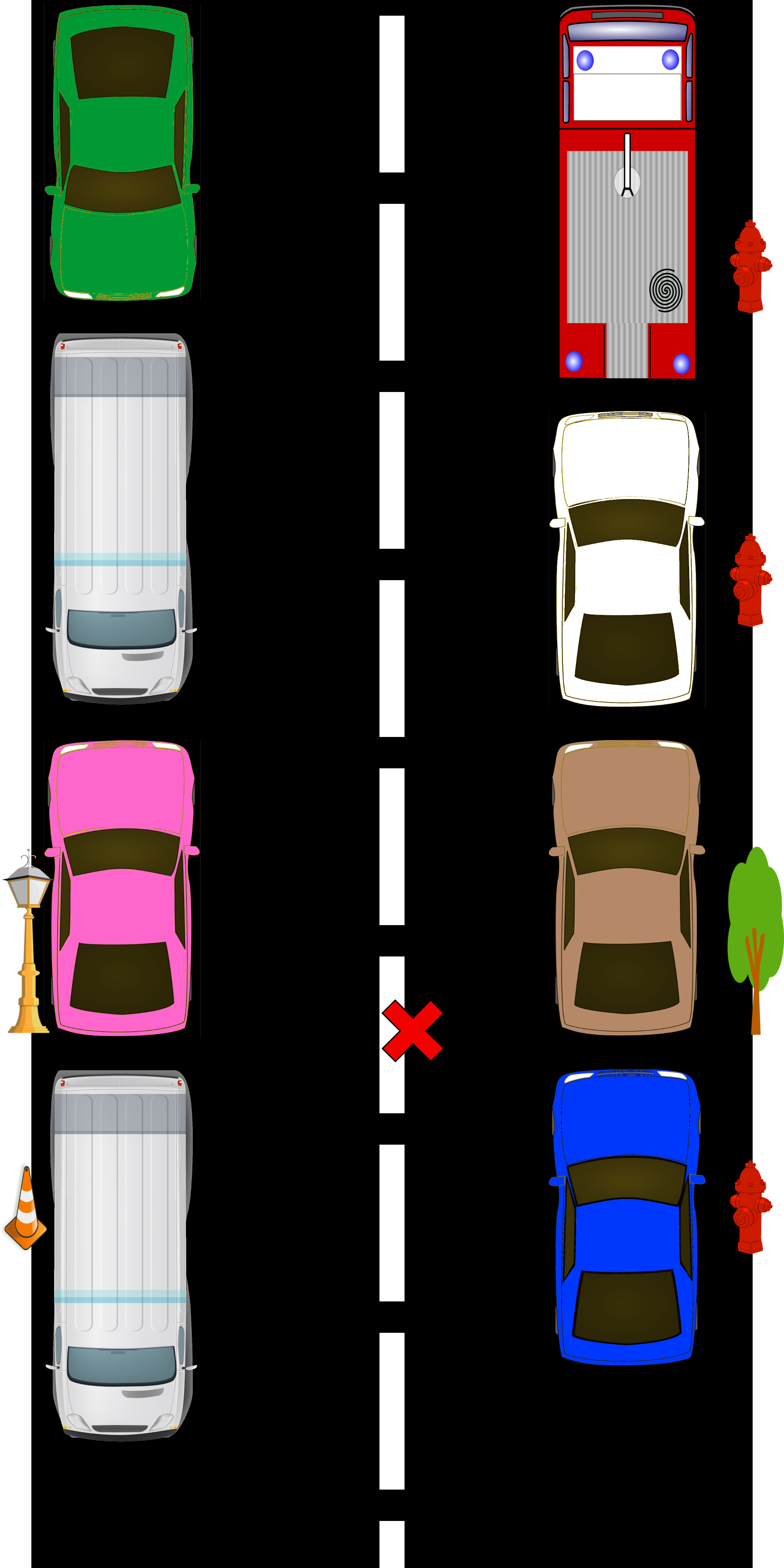}
    \includegraphics[width=0.65\columnwidth]{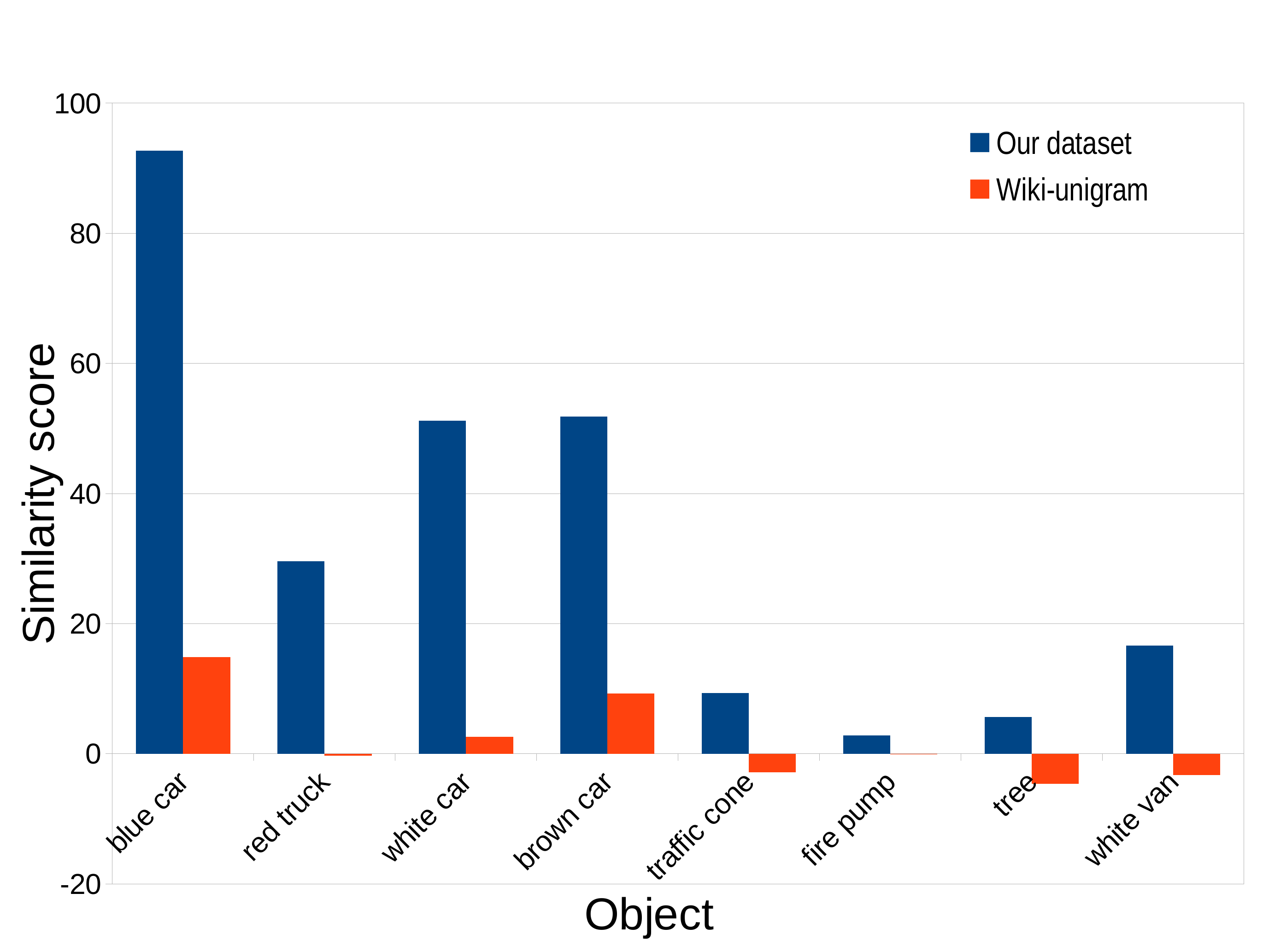} 
    \caption{Street image as seen by the user. For the description of ``drop me off in front of the blue car'', the referent (the blue car) is extracted and then the similarity scores are calculated with the objects present in the image. We can see that the model identifies the correct referent using this method. }
    \label{fig:working}
  \end{figure}

\paragraph{Embeddings} To choose the best embeddings we ran experiments with two approaches: i) out-of-the-box (Wiki-unigram embeddings) and ii) embeddings trained on user descriptions from this domain (training set only).
We split the data into 30\% for the testing set and 70\% for the training set. 
Figure~\ref{fig:working} shows the comparison of the similarity score ($\vec{O}. \vec{R_i}$) using the embeddings trained on our corpus (in blue) and the pre-trained vector (in red). 
The embeddings trained on the training set gave a good representation of the similarity scores despite being sparse. 
A major drawback was the limited vocabulary of the training set. 
This means that words present in the test set but absent in the training set are problematic and thus the sentence embeddings for such descriptions are not produced satisfactorily. 
However, the Wiki-unigram embeddings had a much larger vocabulary (1.7 billion words). This larger vocabulary resulted in a better estimate of the vectors for the REF and ground truth object descriptions.  
Hence, to extract the sentence embeddings, we use the pre-trained Wiki-unigram embeddings (600 dim, trained on English Wikipedia). Table~\ref{tab:ref_resol_results} shows the reference resolution accuracy of the model. This method yields the best performance at 70.2\% accuracy in finding the referent.


  \begin{table}[]
      \centering
      \begin{tabular}{| l | l |}
        \hline
          Random baseline  & 6.25\%\\ \hline      
          Sub-string matching  & 47.5\%\\ \hline 
          Embedding model (training set)     & 60\%\\ \hline
          Embedding model (Wiki-unigram)     & 70.2\%\\ \hline
        \end{tabular}
      \caption{Results of reference resolution performed using different methods (synthetic images). We can see that the sentence embedding models outperform the baseline and sub-string matching. The out-of-the-box embedding model performs significantly better than the model trained using the in-domain trained embeddings (p$<$.05). }
      \label{tab:ref_resol_results}
  \end{table}
  

\section{Conclusion and Future Work}

We introduced a novel dataset of users providing instructions about target locations in a taxi-ride scenario. 
We collected the dataset in two parts, with synthetic images and real-world images. 
We showed that the dataset can be used in many challenging tasks: i) visual reference resolution, ii) direction description understanding, and iii) action identification. 
We presented our novel annotation scheme for natural language and image-related information and performed referent identification experiments on the synthetic images data. 

Our approach is still limited in its capability.
Cases where multiple similar objects were present in the image were not well handled. In such cases, a single sentence/phrase description may not be sufficient to estimate the referent, and we believe that a conversation between the driver and the rider could clarify the referent. 
We will extend our work to include dialogue data between the driver and the rider in a similar simulated setting. 
Our model is currently not capable of performing the reference resolution of objects when multiple similar objects are present in the scene and the user description is sufficient to resolve the references, e.g., ``the second blue truck on the right'', ``the last car on the left'', etc. 
Another case where the model fails to perform well is with plural descriptions of the referents (e.g., ``park in between the 2 blue cars''). 
In such cases we resolve the tie by randomly selecting one of the objects as the referent. We intend to address these issues in future work.

We also intend to validate and extend this work to real-world images. Note that the real-world images descriptions contain more elaborate referent descriptions with e.g., names of car brands, sticker on the car, which can further complicate the task. 

Our annotation scheme has been developed to be task specific. Investigating whether the ISO-Space annotation framework \cite{pustejovsky2011using,pustejovsky2014image} can be applied to our domain is a fruitful direction for future work. 





\section*{Acknowledgments}
This work was partially
supported by the U.S. Army; statements and opinions expressed do not necessarily reflect the position or policy
of the U.S.\ Government, and no official endorsement should be inferred.


\bibliography{references}
\bibliographystyle{acl}

\end{document}